\documentclass[10pt,twocolumn,letterpaper]{article}

\usepackage[pagenumbers]{cvpr} 
\usepackage{multirow}
\usepackage{bbding}
\usepackage{setspace}

\definecolor{cvprblue}{rgb}{0.21,0.49,0.74}
\usepackage[pagebackref,breaklinks,colorlinks,allcolors=cvprblue]{hyperref}

\definecolor{MyGreen}{RGB}{120, 200, 0} 

\title{STCDiT: Spatio-Temporally Consistent Diffusion Transformer for High-Quality Video Super-Resolution}

\author{Junyang Chen \quad Jiangxin Dong$^{\dagger}$ \quad Long Sun \quad Yixin Yang \quad Jinshan Pan$^{\dagger}$ \quad \\
School of Computer Science and Engineering, Nanjing University of Science and Technology\\
{\tt \url{https://jychen9811.github.io/STCDiT_page}}
}

\begin{document}
\maketitle

\begin{abstract} 
We present STCDiT, a video super-resolution framework built upon a pre-trained video diffusion model, aiming to restore structurally faithful and temporally stable videos from degraded inputs, even under complex camera motions.
The main challenges lie in maintaining temporal stability during reconstruction and preserving structural fidelity during generation.
To address these challenges, we first develop a motion-aware VAE reconstruction method that performs segment-wise reconstruction, with each segment clip exhibiting uniform motion characteristic, thereby effectively handling videos with complex camera motions.
Moreover, we observe that the first-frame latent extracted by the VAE encoder in each clip, termed the anchor-frame latent, remains unaffected by temporal compression and retains richer spatial structural information than subsequent frame latents.
We further develop an anchor-frame guidance approach that leverages structural information from anchor frames to constrain the generation process and improve structural fidelity of video features.
Coupling these two designs enables the video diffusion model to achieve high-quality video super-resolution.
Extensive experiments show that STCDiT outperforms state-of-the-art methods in terms of structural fidelity and temporal consistency.

\end{abstract}

\begin{figure*}[t]
	\begin{center}
		\includegraphics[width=0.98\linewidth]{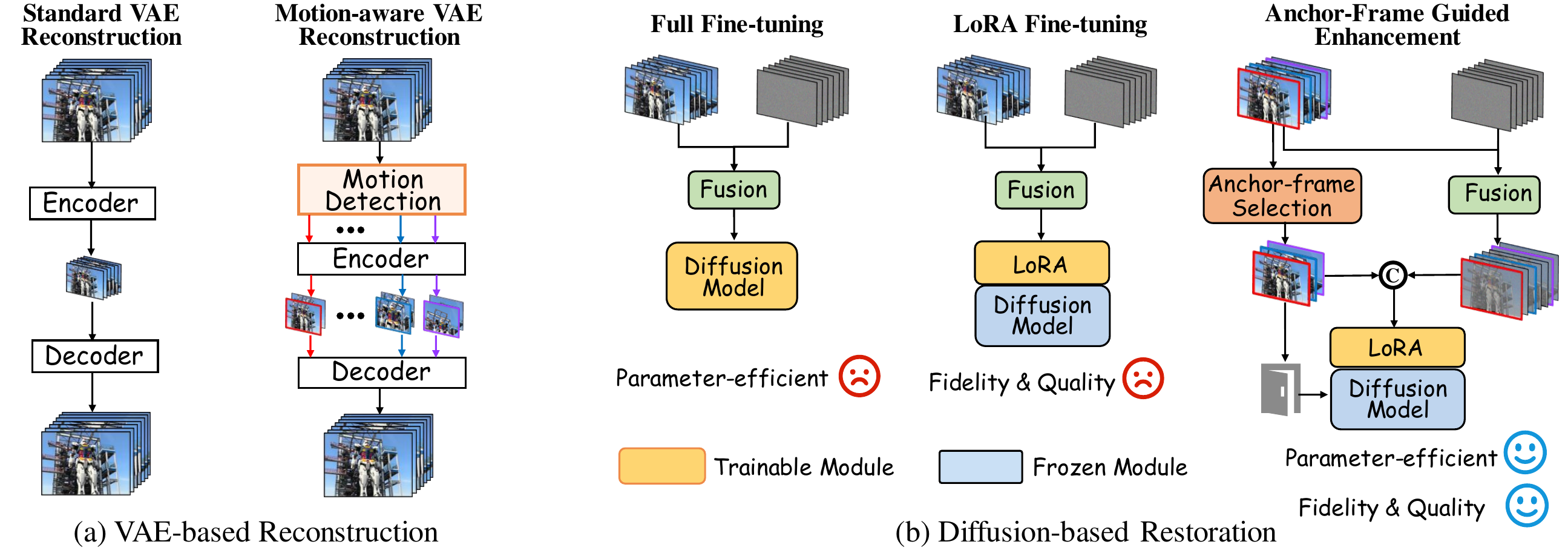}
	\end{center}
        \vspace{-6mm}
	\caption{\textbf{Comparison of VAE-based reconstruction (a) and diffusion-based restoration (b).} Motion-aware reconstruction in (a) allows VAE to handle videos with complex camera motions.  Anchor-frame guided enhancement in (b) enables the parameter-efficient fine-tuning of diffusion transformers for better restoration.
    }
	\label{fig: introduction}
	\vspace{-7mm}
\end{figure*}

\section{Introduction}

Video super-resolution (VSR) aims at restoring high-quality (HQ) video frames from low-quality (LQ) video sequences.
Previous approaches~\cite{recurrent_based_1,basicvsr,basicvsr++} have advanced VSR by exploring spatio-temporal dependencies, leading to significant progress.
However, these methods often fall short in restoring realistic details from degraded inputs.
Diffusion-based techniques offer a promising alternative, owing to their powerful generative priors to address this limitation.
Several recent methods~\cite{UAV, MGLD, DiffVSR} employ image diffusion models to estimate latent high-quality frames, yielding superior perceptual quality.
However, such methods struggle to maintain temporal consistency, as image diffusion models are limited in generating coherent frame sequences with stable temporal consistency.
Although video diffusion models are able to model spatio-temporal information and generate videos with improved temporal coherence compared to image-based counterparts, their direct application to VSR remains non-trivial and poses significant challenges.

The first challenge stems from \textbf{\emph{maintaining temporal stability during video reconstruction}}.
Most state-of-the-art video diffusion methods~\cite{Cogvideox,Wan,Seedvr} employ pre-trained VAEs for temporal downscaling and upscaling in the reconstruction stage.
However, their temporal scaling operators operate locally along the spatial dimension, thereby limiting their capability to model complex spatial transformations across frames.
This limitation becomes particularly problematic in VSR scenarios, where diverse camera motions (\eg, shaking or scaling) introduce intricate motion patterns.
Thus, directly applying pre-trained VAEs for video reconstruction in VSR~\cite{Star, DOVE, Seedvr} may lead to inaccurate results. 
Considering that redesigning these operators or the VAE architecture is highly labor-intensive, we develop an effective motion-aware VAE reconstruction method without incurring substantial manual or computational costs.
Specifically, we detect diverse motion patterns in a video and segment it into several clips with uniform motion patterns for VAE reconstruction (Figure~\ref{fig: introduction}(a)). 
By integrating this motion-aware approach, the VAE can handle varied motion scenarios and reconstruct spatio-temporally coherent results in VSR, as demonstrated in Section~\ref{ssec:effect of motion aware vae}.

Beyond enhancing the reconstruction performance of the VAE, the second challenge lies in \textbf{\emph{maintaining structural fidelity during the generation process}}.
Existing method~\cite{Seedvr} deploys diffusion transformer (DiT) for VSR by fully fine-tuning the model to achieve faithful restoration results.
However, it demands high computational resources to update all model parameters, hindering the effective use of large-scale DiT capability.
LoRA fine-tuning provides a parameter-efficient method for adapting DiT to VSR tasks.
However, several methods~\cite{Lora_1,Lora_2,Lora_3} show that its inherent low-rank constraint hinders the model's ability to capture complex feature interactions, which may limit its effectiveness in VSR task that requires preserving LQ structural fidelity while producing accurate details (see Section~\ref{ssec:effect of AFGE}).
Thus, merely employing LoRA fine-tuning is inadequate to enforce faithful restoration.

To address this, we propose an effective yet parameter-efficient approach that better leverages structural information from LQ inputs to guide generation process.
We observe that when our motion-aware VAE encodes a sequence of video frames, the first frame in each clip, which is unaffected by temporal compression, retains richer spatial structural information compared to subsequent frames.
In our framework, these first-frame latents, termed anchor-frame latents, capture essential structural variations across the video.
We then propose an anchor-frame guidance approach that leverages the wealth of information from the anchor frames to enhance structural fidelity of the video features throughout the generation process (Figure~\ref{fig: introduction}(b)).
Specifically, we first integrate anchor-frame features with video features for interaction via self-attention, enabling the structural information to propagate across video frames.
To reinforce the structural consistency between anchor-frame features and their corresponding video-frame features, we further develop an anchor-corresponding feature modulation module that selectively extracts useful spatial information from anchor frames and then fuses it with the corresponding frame features in the video.
Taken together, our approach is able to employ DiT for high-quality VSR while increasing training parameters by only 7\% of the LoRA parameters with a rank of 128, as exemplified by Wan-14B~\cite{Wan} model.

The main contributions are summarized as follows:
\emph{(i)} We propose a motion-aware VAE reconstruction method, which does not bring computational burdens but significantly improves the reconstruction performance of VAEs for videos exhibiting diverse and complex camera motions.
\emph{(ii)} We develop an effective yet parameter-efficient anchor-frame guidance approach that better constrains the generation process toward structurally faithful VSR.
\emph{(iii)} We design an anchor-corresponding feature modulation module, a simple yet effective component that selectively aggregates useful spatial features to further improve reconstruction fidelity.
\emph{(iv)} Quantitative and qualitative evaluations demonstrate that our approach outperforms state-of-the-art ones, excelling in both structural fidelity and detail generation.

\section{Related Work}
\subsection{Non-diffusion based video super-resolution}

Previous video restoration methods can be roughly categorized into sliding window-based methods~\cite{window_based_1,window_based_2,window_based_3,window_based_4} and recurrent approaches~\cite{recurrent_based_1,recurrent_based_2,recurrent_based_3, recurrent_based_5, recurrent_based_6, recurrent_based_7}.
The former typically splits a video into multiple clips and mainly focuses on reconstructing the center frame of each clip.
The latter exploits temporal dependencies through recurrent networks.
Sajjadi et al.~\cite{recurrent_based_1} utilize forward propagation to transfer information from preceding frames for restoring the current one.
Chan et al.~\cite{basicvsr} further introduce bidirectional propagation, aggregating information from both forward and backward directions to aid the restoration of the current frame.
It is later extended to grid propagation \cite{basicvsr++}, enabling more flexible feature aggregation from diverse spatiotemporal locations.
Despite these advancements, such methods struggle to restore vivid details in real-world scenarios.
To alleviate this problem, several methods \cite{RealBasicVSR, GAN_based_1, GAN_based_2, GAN_based_3} adopt GAN-based training strategies to improve perceptual quality.
However, in the absence of strong priors, the methods that rely solely on the input video often fail to generate realistic content when the degradation is severe.

\subsection{Diffusion based video super-resolution}

Recent diffusion-based image super-resolution methods~\cite{SeeSR,faithdiff,PASD,Dit4SR,omnissr,SUPIR} have significantly advanced this field.
Several methods~\cite{UAV, DiffVSR} extend image diffusion models to video by incorporating temporal attention, enabling more effective aggregation of temporal information.
Yang et al.~\cite{MGLD} use optical flow for motion compensation, improving temporal alignment in video restoration and mitigating frame flickering compared to image SR methods.
However, image diffusion models may produce varying outputs for the same input under different noise conditions, making it difficult to maintain temporal consistency in videos.
Video diffusion models provide a promising solution to address this instability. 
Some methods~\cite{Star,VividVR} employ video diffusion models for VSR using ControlNet-based paradigms.
However, these methods suffer from insufficient interaction~\cite{Dit4SR} between the LQ information and the generative process, compromising restoration fidelity.
Wang et al.~\cite{Seedvr} attempt to enhance this interaction by concatenating LQ video latents with noise latents and fine-tuning the model with a shifted-window attention mechanism for VSR, but this approach incurs high computational costs.
To overcome this challenge, we propose an effective yet parameter-efficient approach that integrates LoRA fine-tuning with a video diffusion model for VSR.

\section{Proposed Method}

Our goal is to develop an effective framework built upon the video diffusion model for high-quality VSR.
We first propose a motion-aware reconstruction method that enables VAE to reconstruct temporally coherent videos, even in the presence of complex inter-frame motions.
To enhance structural fidelity of video features, we propose an anchor-frame guidance approach that propagates representative structural information from anchor frames to facilitate the restoration process.
Furthermore, we develop an anchor-corresponding feature modulation module that extracts and integrates useful spatial information from anchor frames to strengthen the structural fidelity of corresponding frame features in video.
Figure~\ref{fig: method_pipeline} presents the overall architecture of our method.
The details are illustrated as follows.

\subsection{Motion-aware VAE reconstruction}

VAE reconstruction forms the foundational component of the video diffusion pipeline, playing a critical role in achieving high-quality VSR.
However, due to the limited spatial modeling capability of the temporal scaling operators in VAE, the model often fails to accurately characterize large inter-frame motions, which leads to structural distortions or artifacts in videos involving camera shaking or zooming (see Section~\ref{ssec:effect of motion aware vae}).
To address this problem, we propose a motion-aware reconstruction method without introducing excessive burdens.
This approach identifies diverse motion patterns within a video, splits it into multiple clips characterized by consistent motion attributes, and employs VAE-based reconstruction for each segment individually.

\begin{figure*}[t]
	\begin{center}
		\includegraphics[width=1\linewidth]{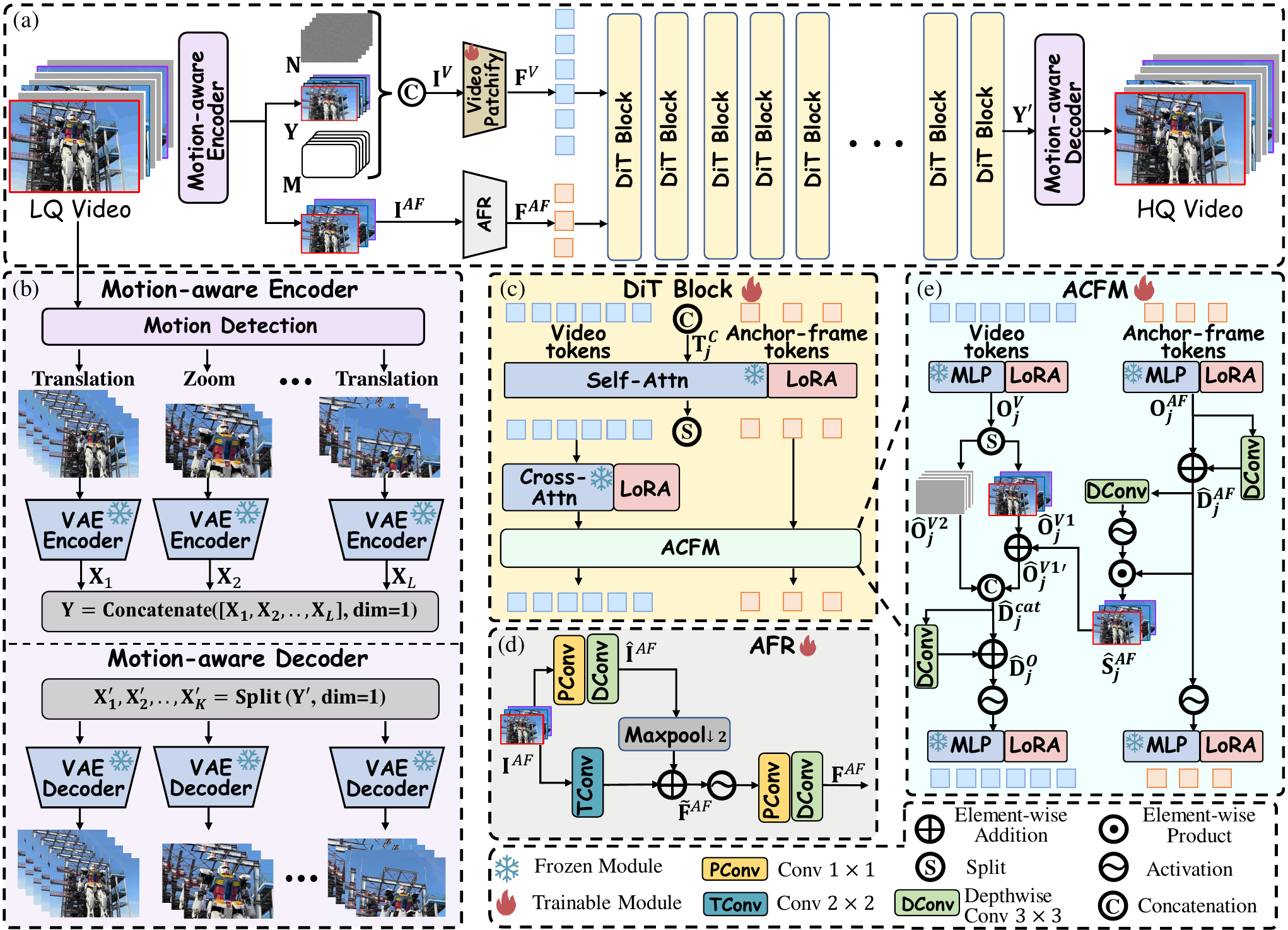}
	\end{center}
        \vspace{-6mm}
	\caption{\textbf{An overview of STCDiT.} For VAE reconstruction, we develop a motion-aware strategy (b) that identifies diverse motion patterns in a video and segments it into clips characterized by consistent motion attributes, which are then independently encoded and decoded by the VAE. For anchor-frame guided enhancement, we sparsely select the first-frame latent from $\{\mathbf{X}_{i}\}^{L}_{i=1}$ as anchor-frame latent and employ a feature refinement module (d) to derive their features. Then, the anchor-frame tokens are concatenated with video tokens, which are then fed into the self-attention layer for feature interaction (c). Moreover, an anchor-frame feature modulation module (e) explores useful information from anchor-frame features, facilitating the restoration of anchor-corresponding frame features in video.
    }
	\label{fig: method_pipeline}
	\vspace{-7mm}
\end{figure*}

To characterize diverse motion patterns in a video, we estimate inter-frame transformation parameters (\emph{i.e.,} translation, rotation angle, and scale) to identify abrupt motion changes.
Specifically, given an LQ video, we first detect corner coordinates across all frames using Shi–Tomasi algorithm~\cite{shi1994good}.
The sparse optical flow estimation (\emph{i.e.,} Lucas–Kanade algorithm~\cite{lucas1981iterative}) is then used to estimate the motion trajectory of the video content, from which an affine transformation matrix is derived.
The transformation parameters are obtained by decomposing the affine transformation matrix. 
Moreover, empirical thresholds are applied to identify abrupt motion changes between frames.
Based on the frame indices where such abrupt transitions occur, the input video is split into $L$ segments. 
Each segment is encoded individually via the VAE, yielding a set of clip latents $\{\mathbf{X}_{i}\}^{L}_{i=1}$, $\mathbf{X}_{i}\in\mathbb{R}^{C \times F \times H \times W}$ (where $C$, $F$, and $H \times W $ denote the channel, temporal, and spatial dimensions).
These latent representations are then concatenated along the temporal dimension to obtain
$\mathbf{Y} \in \mathbb{R}^{C \times F' \times H \times W }$.
Following the diffusion process, the restored video latent $\mathbf{Y'}$ is split according to the original segmentation into $\{\mathbf{X'}_{i}\}^{L}_{i=1}$.
Finally, each $\mathbf{X'}_{i}$ is decoded separately using the VAE to generate the final restored video (see Figure~\ref{fig: method_pipeline}(b)).

\subsection{Anchor-frame guided enhancement}
To better leverage structural information from LQ inputs to guide the generation process, we propose an anchor-frame guided enhancement method.
Specifically, we sparsely select the first frame from $\{\mathbf{X}_{i}\}^{L}_{i=1}$ as anchor-frame latents $\mathbf{I}^{AF}$, which is unaffected by temporal compression and retains rich spatial structural information.
Then, an anchor-frame feature refinement module (Figure~\ref{fig: method_pipeline}(d)) is employed to derive enhanced anchor-frame features $\mathbf{F}^{AF}$ as:
\begin{equation}
    \begin{split}
    & \widehat{\mathbf{I}}^{AF} = \text{DConv}(\text{PConv}(\mathbf{I}^{AF})), \\
    & \widetilde{\mathbf{F}}^{AF} = \downarrow_{\frac{p}{2}}(\widehat{\mathbf{I}}^{AF}) + \text{TConv}(\mathbf{I}^{AF}), \\
    & \mathbf{F}^{AF} = \text{DConv}(\text{PConv}(\zeta(\widetilde{\mathbf{F}}^{AF}))),
    \end{split}
\end{equation}
where $\text{DConv}$ is a $3 \times 3$ depth-wise convolution, $\text{PConv}$ is a $1 \times 1$ convolution, $\downarrow_{\frac{p}{2}}$ is a max-pooling operation with $\times 2$ spatial downsampling, $\zeta$ is the Silu function, and $\text{TConv}$ is a $2 \times 2$ convolution with a stride of 2.
This module enriches the spatial details of the anchor-frame features, thereby facilitating structurally coherent VSR (see Section~\ref{ssec:effect of AFGE}).
Similar to~\cite{Wan}, we obtain the video latent $\mathbf{I}^V$ by concatenating the LQ video latent $\mathbf{Y}$, noise latent $\mathbf{N}$, and an all-one mask $\mathbf{M} \in \mathbb{R}^{C \times F' \times H \times W}$ along the channel dimension. And a `patchify' operation is applied to extract the video features $\mathbf{F}^V$.
Then the video features $\mathbf{F}^V$ and anchor-frame features $\mathbf{F}^{AF}$ are fed into a stack of DiT blocks (Figure~\ref{fig: method_pipeline}(a)).
In the following, we investigate how to effectively leverage structural information from $\mathbf{F}^{AF}$ to further improve the structural fidelity of $\mathbf{F}^{V}$ during the diffusion process.

\noindent \textbf{Anchor-frame guided spatio-temporal feature interaction.}
The input video and anchor-frame features (\ie, $\mathbf{F}^{V}_j$ and $\mathbf{F}^{AF}_j$) to the $j^{th}$ DiT block are first reshaped into token representations and concatenated along the sequence dimension to obtain $\mathbf{T}_j^{C}$, which is then fed into the self-attention layer:
\begin{equation}
    \begin{split}
    & \mathbf{Attn}(\mathbf{T}^{C}_j) = \text{softmax} \left( \frac{\mathbf{Q}_j \mathbf{K}_j ^{\top}}{\sqrt{d}} \right) \mathbf{V}_j,
    \end{split}
\end{equation}
where $\mathbf{Q}_j$, $\mathbf{K}_j$, and $\mathbf{V}_j$ denote query, key, and value matrices obtained from $\mathbf{T}^{C}_j$ through three fully connected layers,
$\text{softmax}(\cdot)$ is a softmax operation, and $d$ is a scaling factor.
Benefiting from the interaction between video features and anchor-frame features in the self-attention layer, video features can leverage structural information embedded in anchor-frame features to enhance their structural fidelity.

Note that for positional encoding of $\mathbf{Q}_j$ and $\mathbf{K}_j$, we preserve the positional indices of video tokens while shifting the indices of anchor-frame tokens along the temporal dimension, which leverages the extrapolation property of RoPE~\cite{su2024roformer} to prevent index overlap between these two token types. 
This strategy enables the pre-trained DiT to incorporate our anchor-frame interaction method without altering the original temporal relationships of the video tokens.

Subsequently, the output of the self-attention layer is split along the sequence dimension to separate video tokens from anchor-frame tokens.
The anchor-frame tokens are excluded from the cross-attention layer, where text embeddings are incorporated, since their interaction could compromise the structural information preserved in the anchor-frame features (Figure~\ref{fig: method_pipeline}(c)).

\noindent \textbf{Anchor-corresponding feature modulation.}
While the feature interaction within the self-attention layer effectively models global dependencies, it is less effective to explore the local spatial information present in the video and anchor-frame features.
To alleviate this problem, we develop an anchor-corresponding feature modulation module (ACFM) inspired by DiT4SR~\cite{Dit4SR}.
Instead of directly injecting anchor-frame features, our module estimates gated units from the anchor-frame features for discriminative feature selection (Figure~\ref{fig: method_pipeline}(e)).

Specifically, we first reshape video tokens and anchor-frame tokens to their corresponding feature representations, denoted as $\mathbf{O}^{V}_j$ and $\mathbf{O}^{AF}_j$, respectively, and extract local information $\mathbf{\widehat{S}}^{AF}_j$ from the anchor-frame features by:
\begin{equation}
    \begin{split}
    & \mathbf{\widehat{D}}^{AF}_j = \text{DConv} ({\mathbf{O}^{AF}_j}) + {\mathbf{O}^{AF}_j}, \\
    & \mathbf{\widehat{S}}^{AF}_j = \mathbf{\widehat{D}}^{AF}_j \odot \phi(\text{DConv}(\mathbf{\widehat{D}}^{AF}_j)),
    \end{split}
\end{equation}
where $\odot$ denotes element-wise product, $\phi$ is the GELU function, and $\text{DConv}$ is a $3 \times 3$ depth-wise convolution.

The extracted local features $\mathbf{\widehat{S}}^{AF}_j$ is subsequently fused with the anchor-corresponding video features $\mathbf{\widehat{O}}^{V1}_{j}$ from $\mathbf{O}^{V}_{j}$ to enhance their structural fidelity, formulated as:
\begin{equation}
    \begin{split}
    & [\mathbf{\widehat{O}}^{V1}_{j}, \mathbf{\widehat{O}}^{V2}_{j}] = \text{Split}(\mathbf{O}^{V}_{j}), \\
    & \mathbf{\widehat{O}}^{V1'}_{j} = \mathbf{\widehat{O}}^{V1}_{j} + \mathbf{\widehat{S}}^{AF}_j, \\
    & \mathbf{\widehat{D}}^{cat}_j = \text{Concat}(\mathbf{\widehat{O}}^{V1'}_{j}, \mathbf{\widehat{O}}^{V2}_{j}), \\
    \end{split}
\end{equation}
where $\text{Split}(\cdot)$ denotes the temporal split operation and $\text{Concat}(\cdot)$ denotes the temporal concatenation operation.

Finally, we apply $\text{DConv}$ to augment local spatial characteristics of the video features $\mathbf{\widehat{D}}^{cat}_j$, formulated as:
\begin{equation}
    \begin{split}
    & \mathbf{\widehat{D}}^{O}_j = \text{DConv}(\mathbf{\widehat{D}}^{cat}_j) + \mathbf{\widehat{D}}^{cat}_j.
    \end{split}
\end{equation}
In contrast to ~\cite{Dit4SR}, our approach discriminatively selects and integrates local spatial features from anchor frames, thereby providing more effective facilitation for VSR.
We will show its effectiveness in Section~\ref{ssec:effect of AFGE}.

\begin{table*}[!t] 
\caption{Quantitative comparisons with state-of-the-art methods on synthetic benchmarks (\ie, REDS30~\cite{REDS} and UDM10~\cite{UDM10}) and real-world benchmarks (\ie, RealVSR~\cite{RealVSR}, VideoLQ~\cite{RealBasicVSR} and SportsLQ). The best and second performances are marked in {\color{red}\textbf{red}} and {\color{blue} blue}, respectively. `–' indicates the method~\cite{MGLD} is not evaluated on REDS30~\cite{REDS}, as a subset of this dataset is used for training.}
\vspace{-3mm}
\renewcommand\arraystretch{1.03}
\normalsize  
\resizebox{\textwidth}{!}{
\begin{tabular}{c|l|ccccccccc}
\toprule
\multirow{2}{*}{Datasets} & \multirow{2}{*}{Metrics}   & {MGLD~\cite{MGLD}} & {UAV~\cite{UAV}} & {STAR~\cite{Star}} & {DOVE~\cite{DOVE}} & SeedVR~\cite{Seedvr}    & SeedVR2~\cite{Seedvr2}     & Wan~\cite{Wan} & {STCDiT-tiny} & {STCDiT}  \\  

&     &  \multicolumn{1}{c}{0.87B}   &  \multicolumn{1}{c}{0.69B}    &  \multicolumn{1}{c}{2B}    &   \multicolumn{1}{c}{5B}      & \multicolumn{1}{c}{7B} & \multicolumn{1}{c}{7B} & \multicolumn{1}{c}{14B}  &   \multicolumn{1}{c}{1.3B}      &     \multicolumn{1}{c}{14B}                           \\

\midrule
           & PSNR {\color{red}$\uparrow$}      & --     & 21.72     & 22.26  & {\color{red}\textbf{23.85}} & 22.34     & 23.17   & 23.30 & {\color{blue} 23.42}  & 23.41 \\
           & SSIM {\color{red}$\uparrow$}      & --     & 0.5328     & 0.5852  & {\color{red}\textbf{0.6411}} & 0.5855     & {\color{blue}0.6338}   & 0.6210  & 0.6304  & 0.6236 \\
           & LPIPS {\color{red}$\downarrow$}   & --     & 0.3659     & 0.4289  & 0.3487 & 0.3209    & 0.3202    & {\color{blue}0.2943} & 0.3398 & {\color{red}\textbf{0.2866}} \\
           & LIQE {\color{red}$\uparrow$}     & --     & 2.217     & 1.573  & 1.7412 & {\color{red}\textbf{2.9496}}     &   2.609  & 2.219  & 2.459 & {\color{blue} 2.875} \\
           & MUSIQ {\color{red}$\uparrow$}     & --     & {\color{blue}59.54}     & 40.45  & 50.33 & 54.50     &   52.17  & 55.05  & 58.96 & {\color{red}\textbf{61.65}} \\
           & CLIPIQA+ {\color{red}$\uparrow$}  & --     & 0.4539     & 0.3158  & 0.3963 & {\color{blue}0.4699}     &  0.4364  & 0.4238 & 0.4573  & {\color{red}\textbf{0.4728}} \\
           & MANIQA {\color{red}$\uparrow$}    & --     & 0.3046     & 0.1824  & 0.2306 & 0.2765     &  0.2596   & 0.2547  & {\color{red}\textbf{0.3271}}  & {\color{blue} 0.3147} \\
           & FasterVQA {\color{red}$\uparrow$}    & --     & 0.6598     & 0.5884  & 0.6534 & 0.6425     &  0.6062   & {\color{blue}0.7144}  & 0.6956  & {\color{red}\textbf{0.7338}} \\
\multirow{-9}{*}{REDS30~\cite{REDS}}    &   DOVER {\color{red}$\uparrow$}    & --     & 32.23     & 36.43  & 36.83 & 36.36     & 35.54    & 39.69  & {\color{blue} 40.09} & {\color{red}\textbf{42.94}} \\ \midrule

           & PSNR {\color{red}$\uparrow$}     & 28.11     & 26.55  & 25.53  & {\color{red}\textbf{29.39}} & 27.07     & 28.29  & {\color{blue}28.64} & 28.59   & 28.37     \\
           & SSIM {\color{red}$\uparrow$}      & 0.8346     & 0.7988      & 0.7899  & {\color{red}\textbf{0.8621}} & 0.8153    & {\color{blue}0.8518}   & 0.8425 & 0.8451  & 0.8354     \\
           & LPIPS {\color{red}$\downarrow$}   & 0.1971     & 0.2320      & 0.2312  & {\color{blue} 0.1581} & 0.1827     & {\color{red}\textbf{0.1524}}  & 0.1720 & 0.1749 & 0.1682     \\
           & LIQE {\color{red}$\uparrow$}    & 3.041     & 2.939     & 2.190  & 2.719 & {\color{blue} 3.234}     & 3.004  & 3.008  & 3.055  & {\color{red}\textbf{3.420}} \\
           & MUSIQ {\color{red}$\uparrow$}    & 62.04     & 63.95     & 60.84  & 60.84 & 64.62     & 63.82  & 63.75 & {\color{blue} 64.83}   & {\color{red}\textbf{66.46}} \\
           & CLIPIQA+ {\color{red}$\uparrow$}  & 0.4017     & 0.4843      & 0.4296  & 0.4296 & 0.4722     & 0.4542  & 0.4918 & {\color{blue} 0.5007 }  & {\color{red}\textbf{0.5234}} \\
           & MANIQA {\color{red}$\uparrow$}    & 0.2992     & 0.3479     & 0.2649  & 0.3086 & 0.3413     & 0.3327   & 0.3547  & {\color{blue} 0.3573} & {\color{red}\textbf{0.3763}} \\
           & FasterVQA {\color{red}$\uparrow$}    & 0.7861     & 0.8058     & 0.7813  & 0.7954 & 0.8140     &  0.8012   & 0.8086  & {\color{blue}0.8157}  & {\color{red}\textbf{0.8317}} \\
\multirow{-9}{*}{UDM10~\cite{UDM10}}       & DOVER {\color{red}$\uparrow$}      & 57.78     & 56.48     & 54.03  & 57.70 & 60.93     &  60.59   & 60.64 & {\color{blue} 61.08}  & {\color{red}\textbf{64.10}} \\ \midrule

           & PSNR {\color{red}$\uparrow$}      & 22.06     & 20.91  & 17.82  & {\color{blue}22.37} & 21.87     & 22.10  & {\color{red}\textbf{22.47}}  & 22.04  & 21.51     \\
           & SSIM {\color{red}$\uparrow$}      & 0.6773     & 0.6078      & 0.5362  & {\color{blue} 0.7318} & 0.6952    & 0.7215  & {\color{red}\textbf{0.7441}}  & 0.7086  & 0.7263     \\
           & LPIPS {\color{red}$\downarrow$}   & 0.2031     & 0.2513      & 0.2846  & 0.1766 & 0.1829     & 0.1828   & {\color{blue} 0.1655}  & 0.1666  & {\color{red}\textbf{0.1553 }}   \\
           & LIQE {\color{red}$\uparrow$}      &  {\color{red}\textbf{4.225}}     & 4.081     & 3.619     & 3.870     & 3.6403     & 3.482  & 3.891   & 4.019 &  {\color{blue} 4.145}     \\
           & MUSIQ {\color{red}$\uparrow$}     & 70.00     & 69.85      & 71.07  & 71.29 & 65.12      & 63.52  & 70.32  & {\color{blue} 72.82}  & {\color{red}\textbf{73.88}} \\
           & CLIPIQA+ {\color{red}$\uparrow$}  & 0.4710     & 0.5194     & 0.4958  & 0.4908 & 0.4823     & 0.4643  & 0.5059  & {\color{red}\textbf{0.5616}}   & {\color{blue} 0.5393} \\
           & MANIQA {\color{red}$\uparrow$}    & 0.3786     & {\color{blue}0.4240}     &  0.3919  & 0.3756 &  0.3411     & 0.3370   & 0.3844  & 0.4142  & {\color{red}\textbf{0.4338}} \\
           & FasterVQA {\color{red}$\uparrow$}    & 0.7764     & 0.7449     & 0.7672  & {\color{blue}0.8024} & 0.7657     &  0.7416   & 0.7882  & 0.7940  & {\color{red}\textbf{0.8031}} \\
\multirow{-9}{*}{RealVSR~\cite{RealVSR}}     & DOVER {\color{red}$\uparrow$}       & 53.37     & 46.62     & 57.53  & {\color{blue} 59.60} & 52.25     & 54.41   & 56.37  & 57.13 & {\color{red}\textbf{61.57}} \\ \midrule

           & LIQE {\color{red}$\uparrow$}          &  1.764     & 1.867     & 1.644     & 1.732     & 1.338     & 1.126  & 1.885 & {\color{blue} 1.9359}   &  {\color{red}\textbf{2.226}}     \\
           & MUSIQ {\color{red}$\uparrow$}          &  47.86    & 47.59     & 42.39     & 44.74     & 40.66     & 35.06  & 45.22 & {\color{blue} 48.18}   &  {\color{red}\textbf{48.54}}     \\
           & CLIPIQA+ {\color{red}$\uparrow$}       & 0.3986     & 0.3929     &  0.4146     & 0.3978     & {\color{red}\textbf{0.4225} }    & 0.3789   & 0.3862 & 0.4103  &  {\color{blue} 0.4207}    \\ 
           & MANIQA {\color{red}$\uparrow$}         & 0.2572     &  0.2704     & 0.2524     & 0.2576     & 0.2294     & 0.2153  & 0.2700  & {\color{blue} 0.2974}  &  {\color{red}\textbf{0.2990}}      \\
           & FasterVQA {\color{red}$\uparrow$}    & 0.7283     & 0.7178     & 0.6886  & 0.7107 & 0.6342     &  0.5201   & 0.7346 & {\color{blue}0.7420}   & {\color{red}\textbf{0.7596}} \\
\multirow{-6}{*}{VideoLQ~\cite{RealBasicVSR}}     & DOVER {\color{red}$\uparrow$}      & 51.49     & 48.27     & 50.34     &   53.67     & 45.64  & 42.57  & 53.69  & {\color{blue}54.02}  &  {\color{red}\textbf{57.01}}     \\ \midrule

           & LIQE {\color{red}$\uparrow$}   & 1.986     & 1.555     & 1.459     & 1.991     & 1.289     & 1.316     & 1.754  & {\color{blue} 1.999 }  & {\color{red}\textbf{2.099}}      \\
           & MUSIQ {\color{red}$\uparrow$}   & 55.66     & 49.13     & 53.47     & 56.32     & 44.37     & 46.36   & 52.13  & {\color{blue} 56.01}     & {\color{red}\textbf{57.51}}      \\
           & CLIPIQA+ {\color{red}$\uparrow$}& 0.3310     & 0.3686     & 0.4005     & 0.4014     & 0.3423     & 0.3578     & 0.4110  & {\color{red}\textbf{0.4472}}  & {\color{blue} 0.4393 }     \\
           & MANIQA {\color{red}$\uparrow$}  & 0.2465     & 0.2031     & 0.2274     & 0.2594     & 0.1725     & 0.1881     & 0.2338  & {\color{blue} 0.2666}   & {\color{red}\textbf{0.2682 }}     \\
           & FasterVQA {\color{red}$\uparrow$}  & 0.7364     & 0.5928     & 0.6749     & 0.7332     & 0.5218     & 0.5274     & 0.6946  &  {\color{blue} 0.7446}   & {\color{red}\textbf{0.7574}}     \\
\multirow{-6}{*}{SportsLQ}       & DOVER {\color{red}$\uparrow$}    & 42.73     & 33.22     & 46.01     & 44.91     & 43.76     & 33.51     & 44.04  & {\color{blue} 46.42 }   & {\color{red}\textbf{48.98}}    \\  
\bottomrule
\end{tabular}}
\vspace{-6mm}
\label{tab:quantitative-results}
\end{table*}

\section{Experimental Results}
\label{sec: Experiment}
We evaluate the proposed approach against state-of-the-art ones on publicly available benchmark datasets. More experimental results are included in the supplemental material.

\subsection{Datasets and implementation details}
\noindent \textbf{Datasets.} 
We combine HQ videos from UltraVideo~\cite{UltraVideo} and HQ images from LSDIR~\cite{LSDIR} as our training data.
The corresponding LQ counterparts are synthesized by applying the degradation pipelines from RealBasicVSR~\cite{RealBasicVSR} and Real-ESRGAN~\cite{Real-ESRGAN} to the HQ video and image data, respectively.
To further enhance the realism of the degradations, we incorporate camera shake and scaling effects.
Textural captions for the training data are generated using Qwen2.5-VL~\cite{QwenVL}.
We evaluate our approach on four commonly used testing datasets, including two synthetic datasets (\ie, REDS30~\cite{REDS} and UDM10~\cite{UDM10}) and two real-world datasets (\ie, RealVSR~\cite{RealVSR} and VideoLQ~\cite{RealBasicVSR}).
Additionally, we introduce a new real-world dataset, SportsLQ, which consists of 20 sports event videos at 720p resolution, each containing 100 frames, to complement existing benchmarks.
All experiments are conducted with a $\times 4$ upscaling factor, except for evaluations on RealVSR~\cite{RealVSR} and SportsLQ, which are performed at the original scale ($\times 1$).

\begin{figure*}[!t]
\footnotesize
\centering
    \begin{tabular}{c c c c c c c}

            \multicolumn{3}{c}{\multirow{5}*[83.1pt]{
            \hspace{-4mm} \includegraphics[width=0.255\linewidth, height=0.389\linewidth]{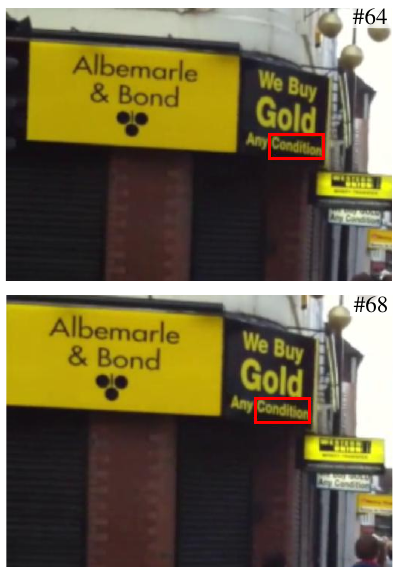}}}
            & \hspace{-5.1mm} \includegraphics[width=0.185\linewidth,height=0.18\linewidth]{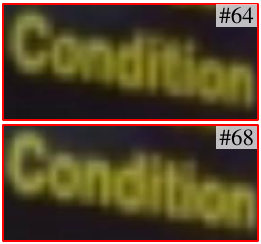}
            & \hspace{-4.6mm} \includegraphics[width=0.185\linewidth,height=0.18\linewidth]{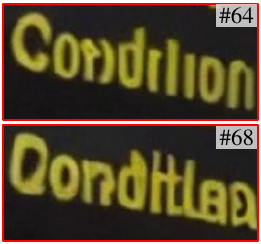} 
            & \hspace{-4.6mm} \includegraphics[width=0.185\linewidth,height=0.18\linewidth]{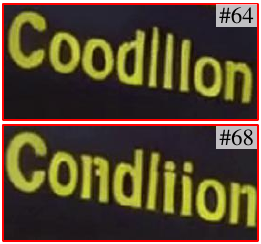} 
            & \hspace{-4.6mm} \includegraphics[width=0.185\linewidth,height=0.18\linewidth]{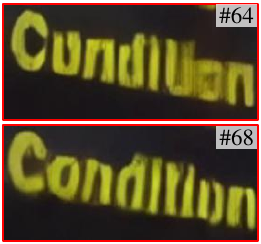} 
              \\
    		\multicolumn{3}{c}{~}                                   
            & \hspace{-5.1mm} (a1) Bicubic 
            & \hspace{-4.6mm} (b1) UAV~\cite{UAV}
            & \hspace{-4.6mm} (c1) STAR~\cite{Star}
            & \hspace{-4.6mm} (d1) DOVE~\cite{DOVE} \\		
    	\multicolumn{3}{c}{~} 
            & \hspace{-5.1mm} \includegraphics[width=0.185\linewidth,height=0.18\linewidth]{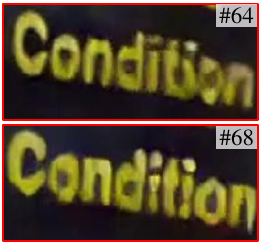} 
            & \hspace{-4.6mm} \includegraphics[width=0.185\linewidth,height=0.18\linewidth]{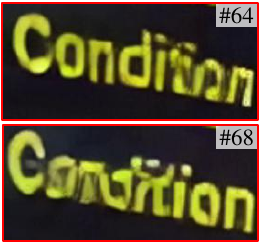} 
            & \hspace{-4.6mm} \includegraphics[width=0.185\linewidth,height=0.18\linewidth]{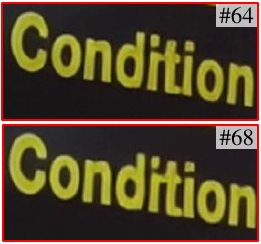} 
            & \hspace{-4.6mm} \includegraphics[width=0.185\linewidth,height=0.18\linewidth]{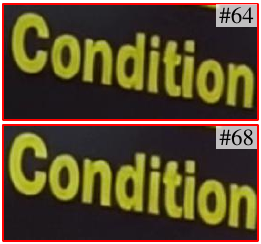} 
            \\
    	\multicolumn{3}{c}{\hspace{-4.0mm} LQ video 013 from VideoLQ~\cite{RealBasicVSR}}
            & \hspace{-5.1mm} (e1) SeedVR-7B~\cite{Seedvr}
            & \hspace{-4.6mm} (f1) Wan~\cite{Wan}
            & \hspace{-4.6mm} (g1) STCDiT-tiny
            & \hspace{-4.6mm} (h1) STCDiT\\

            \multicolumn{3}{c}{\multirow{5}*[83.6pt]{
            \hspace{-4.5mm} \includegraphics[width=0.255\linewidth, height=0.390\linewidth]{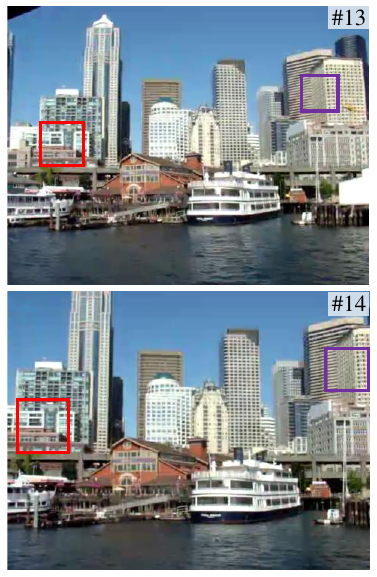}}}
            & \hspace{-5.1mm} \includegraphics[width=0.185\linewidth,height=0.18\linewidth]{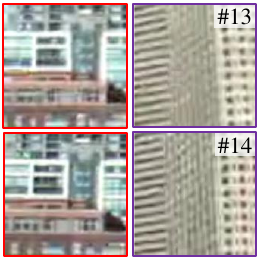}
            & \hspace{-4.6mm} \includegraphics[width=0.185\linewidth,height=0.18\linewidth]{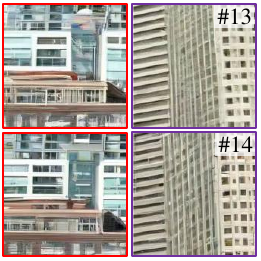} 
            & \hspace{-4.6mm} \includegraphics[width=0.185\linewidth,height=0.18\linewidth]{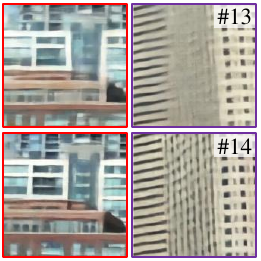} 
            & \hspace{-4.6mm} \includegraphics[width=0.185\linewidth,height=0.18\linewidth]{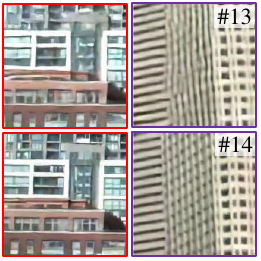} 
              \\
    		\multicolumn{3}{c}{~}                                   
            & \hspace{-5.1mm} (a2) Bicubic 
            & \hspace{-4.6mm} (b2) MGLD~\cite{MGLD}
            & \hspace{-4.6mm} (c2) DOVE~\cite{DOVE}
            & \hspace{-4.6mm} (d2) SeedVR-7B~\cite{Seedvr} \\		
    	\multicolumn{3}{c}{~} 
            & \hspace{-5.1mm} \includegraphics[width=0.185\linewidth,height=0.18\linewidth]{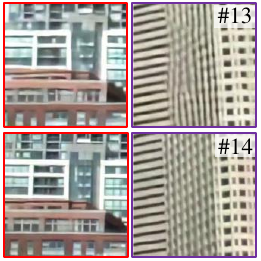} 
            & \hspace{-4.6mm} \includegraphics[width=0.185\linewidth,height=0.18\linewidth]{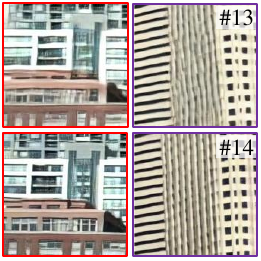} 
            & \hspace{-4.6mm} \includegraphics[width=0.185\linewidth,height=0.18\linewidth]{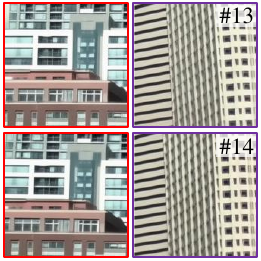} 
            & \hspace{-4.6mm} \includegraphics[width=0.185\linewidth,height=0.18\linewidth]{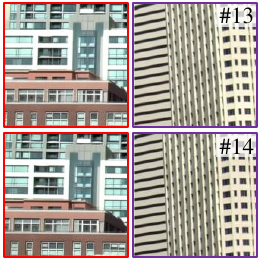} 
            \\
    	\multicolumn{3}{c}{\hspace{-4.0mm} LQ video 024 from VideoLQ~\cite{RealBasicVSR}}
            & \hspace{-5.1mm} (e2) SeedVR2-7B~\cite{Seedvr2}
            & \hspace{-4.6mm} (f2) Wan~\cite{Wan}
            & \hspace{-4.6mm} (g2) STCDiT-tiny
            & \hspace{-4.6mm} (h2) STCDiT\\

    \end{tabular}
\vspace{-3mm}
\caption{VSR results ($\times 4$) on the real-world benchmark. We provide two types of video motion scenarios: camera shaking and camera zooming. The prompt corresponding to the video of the first row does not contain the word `condition'. Compared to competing methods, our method restores temporally consistent videos with fidelity structure and vivid details, even under complex inter-frame motions. }
\label{fig: Results}
\vspace{-6mm}
\end{figure*}

\noindent \textbf{Implementation Details.} 
We implement our method based on Wan2.1 T2V-1.3B and Wan 2.1 I2V-14B~\cite{Wan}, denoted as `STCDiT-tiny' and `STCDiT', respectively.
The LoRA rank is set to $128$ in our experiment.
We randomly crop video patches of spatial size $480 \times 480$ with a temporal length of 27-33 frames and image patches of spatial size $512 \times 512$ as the training inputs.
The batch sizes of video and image are set to $32$ and $128$.
We use AdamW optimizer with a constant learning rate $5e-5$.
We use mean squared error loss to constrain the model training, similar to~\cite{Wan, Flow_matching}.
We conduct our experiments using 4 NVIDIA A800 GPUs.
During the inference stage, the inference step is set to $10$.

\subsection{Quantitative results}

To evaluate the restoration performance of the proposed method, we compare it with state-of-the-art methods (\ie, MGLD~\cite{MGLD}, UAV~\cite{UAV}, STAR-I2VGEN~\cite{Star}, DOVE~\cite{DOVE}, SeedVR-7B~\cite{Seedvr} and SeedVR2-7B~\cite{Seedvr2}). 
We also include a comparison with Wan 2.1 I2V-14B~\cite{Wan}, which employs a LoRA-based training strategy under the same experimental settings as ours, denoted as Wan.
We employ no-reference metrics (\ie, LIQE~\cite{LIQE}, NIQE~\cite{NIQE}, MUSIQ~\cite{MUSIQ}, CLIPIQA+~\cite{CLIPIQA}, MANIQA~\cite{MANIQA}, FasterVQA~\cite{FasterVQA} and DOVER~\cite{DOVER}) to evaluate restoration quality across both synthetic and real-world datasets while applying reference-based metrics (\ie, PSNR, SSIM, and LPIPS~\cite{Lpips}) to assess fidelity on datasets with high-quality ground-truth pairs.
We first evaluate our method on REDS30~\cite{REDS}, UDM10~\cite{UDM10} and RealVSR~\cite{RealVSR}.
Table~\ref{tab:quantitative-results} shows the quantitative results.
Our STCDiT achieves the best LPIPS values on REDS30~\cite{REDS} and RealVSR~\cite{RealVSR}.
The proposed method also outperforms all competing approaches in terms of MUSIQ, CLIPIQA+, MANIQA, FasterVQA, and DOVER on REDS30~\cite{REDS}, UDM10~\cite{UDM10} and RealVSR~\cite{RealVSR}, demonstrating its effectiveness in handling challenging motions.
Then, we evaluate our method on VideoLQ~\cite{RealBasicVSR}.
It covers a wide range of scenes and diverse video motions such as shaking, zooming, and zooming out, making it challenging to restore high-quality results from such cases.
Nevertheless, our method achieves the best results on almost all metrics.
Finally, we evaluate our method on SportsLQ, including videos with intense human activities.
The proposed method outperforms competing methods in terms of all no-reference metrics.
In addition, our STCDiT-tiny also achieves competitive performance compared to other methods, demonstrating the effectiveness of our approach.

\subsection{Qualitative results}

The first example of Figure~\ref{fig: Results} presents a case with camera shake.
The result obtained by~\cite{UAV} illustrates that merely combining optical flow with an image diffusion model is insufficient for ensuring temporal stability.
STAR~\cite{Star} produces results with structural distortions, as insufficient interaction between the LQ input and the generation process limits the powerful generative capabilities of the pre-trained diffusion model.
The results of~\cite{DOVE, Seedvr, Wan} exhibit structural distortions and artifacts.
This is mainly because these methods directly employ the pre-trained VAE for video reconstruction, where its temporal scaling operators limit its ability to model accurate spatio-temporal information under complex inter-frame motions, thus compromising the VSR quality.
In contrast, by employing the motion-aware VAE reconstruction and leveraging rich structural information from anchor frames to constrain the generation process, our method generates temporally coherent results with faithful structural details (\ie, the letters).
The second example of Figure~\ref{fig: Results} shows a zooming scenario.
The competing methods struggle to restore vivid structural details.
However, our STCDiT exhibits excellent performance in recovering high-quality results with better temporal consistency.

\vspace{-2mm}

\section{Analysis and Discussion}
\label{sec: ablation_study}
For the ablation studies in this section, our method and all baselines are built on Wan 2.1 I2V-14B~\cite{Wan}, trained on the same datasets~\cite{UltraVideo,LSDIR} and batch size as in Section~\ref{sec: Experiment}, with half the total training iterations.

\subsection{Effect of motion-aware VAE reconstruction}
\label{ssec:effect of motion aware vae}

The motion-aware reconstruction approach is proposed to improve the VAE's ability to generate temporally coherent video latents, even in the presence of complex inter-frame motions.
We first validate its effect on a video reconstruction task and compare the standard reconstruction method (\emph{ST VAE} for short) with our motion-aware reconstruction method (\emph{MA VAE} for short), both built upon a pretrained VAE~\cite{Wan}.
We compare their reconstruction performance of ground-truth videos on the REDS30~\cite{REDS} dataset, which contains diverse handheld videos exhibiting complex motion and camera shake.
The quantitative results in Table~\ref{tab: VAE_ablation} show that the PSNR of our \textit{$\text{MA VAE}$} is $4.20$dB higher than \textit{$\text{ST VAE}$}. 
The substantial improvement demonstrates that our motion-aware mechanism enables more effective reconstruction of videos with complex motion patterns.

\begin{table}[!t]
\caption{Effectiveness of the proposed motion-aware VAE reconstruction method.}
\vspace{-2mm}
\tiny
\centering
\resizebox{1.0\columnwidth}{!}{
\begin{tabular}{lccc}
\hline
       & ~~~Motion-aware~~~ & ~~~PSNR~~~ & ~~~SSIM~~~ \\ \hline
$\text{ST VAE}$ &  \XSolidBrush                                 & 27.22     &  0.7802    \\
$\text{MA VAE}$ &   \Checkmark                       & \textbf{31.42}     & \textbf{0.8924} \\
\hline         
\end{tabular}}
\vspace{-6mm}
\label{tab: VAE_ablation}
\end{table}

\begin{table*}[!ht]
\centering
\caption{Effectiveness of the anchor-frame guided enhancement. All methods are evaluated on the dataset of ~\cite{RealVSR}.}
\vspace{-2mm}
\footnotesize
\resizebox{\textwidth}{!}{
\begin{tabular}{lcccccccccccccc}
\toprule
\multirow{2}{*}{}          & \multicolumn{2}{c}{Anchor-frame selection}  &  & \multicolumn{2}{c}{Anchor-frame injection}  &  & Anchor-frame  &  & Anchor-frame interaction &  & \multirow{2}{*}{MUSIQ} & \multirow{2}{*}{DOVER} \\ \cline{2-3} \cline{5-6}
                           & First frames                & Uniform sampling                    &  & w/ DW Conv                  & w/ ACFM                      &  &  refinement                               &  &     w/ text embedding                                &  &                        &                        \\ \midrule
$\text{Base}$                & \XSolidBrush & \XSolidBrush &  & \XSolidBrush & \XSolidBrush &  & \XSolidBrush     &  & \XSolidBrush         &  & 68.30                  & 55.62                  \\
$\text{Base}_{\text{w/ FF}}$                 & \Checkmark   & \XSolidBrush &  & \XSolidBrush & \XSolidBrush &  & \XSolidBrush     &  & \XSolidBrush         &  & 70.58                  & 58.68                  \\
$\text{Base}_{\text{w/ FF}\&\text{DWC}}$       & \Checkmark   & \XSolidBrush &  & \Checkmark   & \XSolidBrush &  & \XSolidBrush     &  & \XSolidBrush         &  & 71.87                  & 59.34                  \\
$\text{Base}_{\text{w/ FF}\&\text{ACFM}}$     & \Checkmark   & \XSolidBrush &  & \XSolidBrush & \Checkmark   &  & \XSolidBrush     &  & \XSolidBrush         &  & 73.24                  & 59.99                  \\
$\text{Ours}_{\text{w/ ITE}}$                  & \Checkmark   & \XSolidBrush &  & \XSolidBrush & \Checkmark   &  & \Checkmark       &  & \Checkmark           &  & 68.73                  & 56.15                  \\
$\text{Ours}_{\text{w/o FF}\&\text{w/ US}}$ & \XSolidBrush & \Checkmark   &  & \XSolidBrush & \Checkmark   &  & \Checkmark       &  & \XSolidBrush         &  & 69.72                  & 56.39                  \\
$\text{Ours}$                & \Checkmark   & \XSolidBrush &  & \XSolidBrush & \Checkmark   &  & \Checkmark       &  & \XSolidBrush         &  & \textbf{73.57}         & \textbf{60.81}         \\ \bottomrule
\end{tabular}}
\vspace{-6mm}
\label{tab: ablation_study_all}
\end{table*}

\begin{figure}[!t]
\scriptsize
\vspace{0mm}
\centering
\begin{tabular}
{@{}c@{\hspace{0.1mm}}c@{\hspace{0.1mm}}c@{\hspace{0.1mm}}c@{\hspace{0.1mm}}c@{\hspace{0.1mm}}c@{}}
\includegraphics[width = 0.20\linewidth]{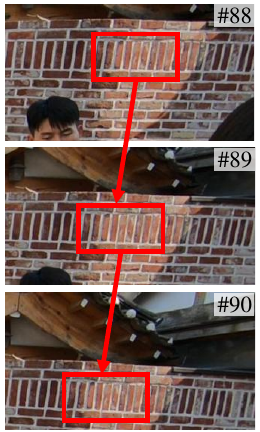}& 
\includegraphics[width = 0.20\linewidth]{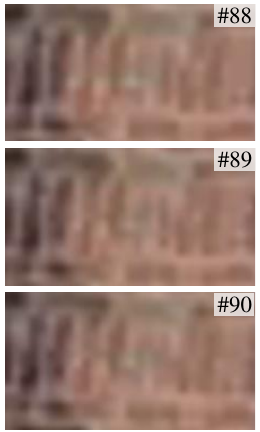}& 
\includegraphics[width = 0.20\linewidth]{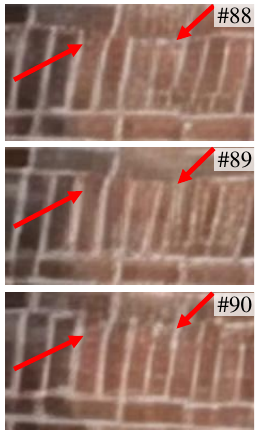}& 
\includegraphics[width = 0.20\linewidth]{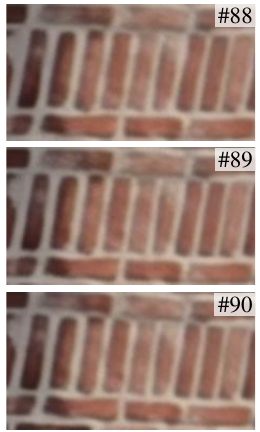} &
\includegraphics[width = 0.20\linewidth]{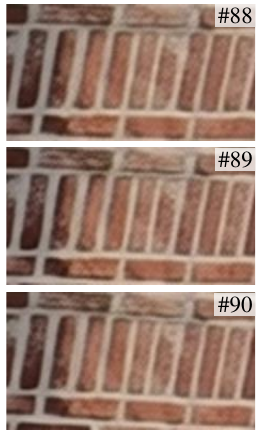} \\
(a) GT Video &\hspace{-0.5mm} 
(b) Bicubic &\hspace{-0.5mm}  
(c) Wan &\hspace{-0.5mm}  
(d) Base &\hspace{-0.5mm} 
(e) Ours \\
\end{tabular}
\vspace{-3.5mm}
\caption{Effectiveness of the motion-aware VAE reconstruction method on a video with camera shaking.}
\label{fig: effectiveness_MA}
\vspace{-6.8mm}
\end{figure}

We further validate the effectiveness of the motion-aware reconstruction method on the restoration task.
We compare Wan~\cite{Wan} with a baseline method that replaces the standard VAE with our motion-aware VAE, whose encoded features are subsequently fed into a LoRA-trained DiT~\cite{Wan} for restoration (\emph{Base} for short).
Figure~\ref{fig: effectiveness_MA} shows a visual comparison on a degraded video with severe camera shaking.
The comparison results demonstrate that the motion-aware VAE reconstruction method is more effective at 
producing temporally coherence video latents (e.g., the bricks in Figure~\ref{fig: effectiveness_MA}(d)).
Also note that our anchor-frame guided enhancement is able to further improve the restoration performance with sharper details (Figure~\ref{fig: effectiveness_MA}(e)).

\subsection{Effect of anchor-frame guided enhancement}
\label{ssec:effect of AFGE}
The proposed anchor-frame guided enhancement utilizes information from LQ inputs to constrain the generation process, promoting structurally faithful VSR.
To demonstrate its effectiveness, we first compare \textit{$\text{Base}$} with a variant that integrates anchor-frame features with video features in the self-attention module (\emph{Base$_{\text{w/ FF}}$} for short).
The results in Table~\ref{tab: ablation_study_all} show that the interaction between video features and anchor-frame features generates better results, improving MUSIQ and DOVER by $2.28$ and $3.06$, respectively.
This demonstrates the importance of the guided interaction between these features in boosting restoration performance.

\begin{figure}[!t]
\scriptsize
\vspace{0mm}
\centering
\tiny
\resizebox{1.0\columnwidth}{!}{
\begin{tabular}
{@{}c@{\hspace{0mm}}c@{\hspace{-0.5mm}}c@{\hspace{-0.5mm}}c@{\hspace{-0.5mm}}c@{}}
\includegraphics[width = 0.185\linewidth]{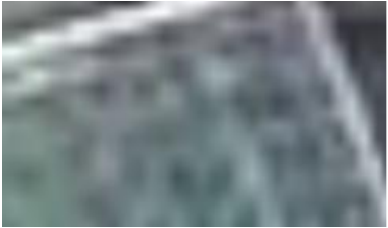}& 
\includegraphics[width = 0.185\linewidth]{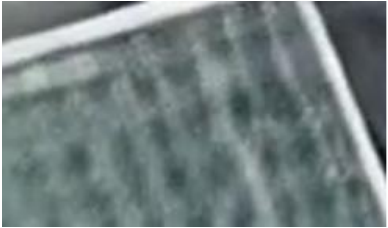}& 
\includegraphics[width = 0.185\linewidth]{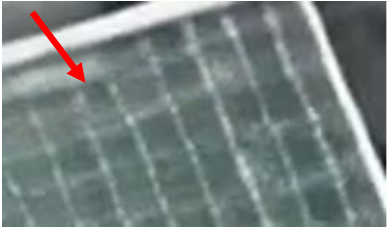}&
\includegraphics[width = 0.185\linewidth]{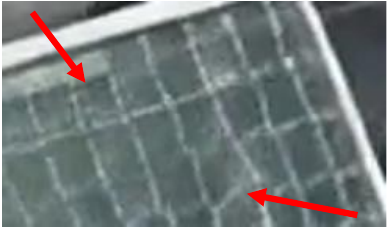} \\
(a) LQ Patch  &\hspace{-0.5mm}  
(b) Base  &\hspace{-0.5mm}  
(c) Base$_{\text{w/ FF}}$  &\hspace{-0.5mm} 
(d) Base$_{\text{w/ FF\&DWC}}$  &\hspace{-0.5mm}   \\
\includegraphics[width = 0.185\linewidth]{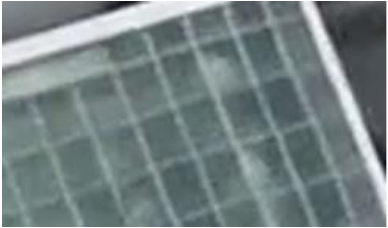}& 
\includegraphics[width = 0.185\linewidth]{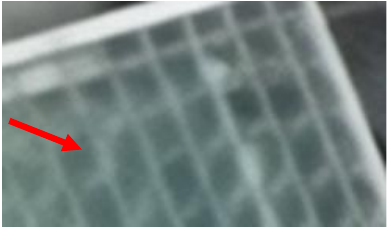}& 
\includegraphics[width = 0.185\linewidth]{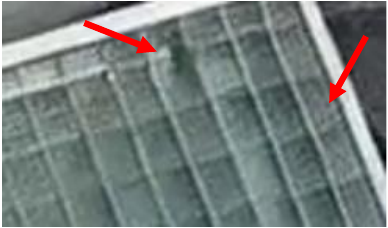}& 
\includegraphics[width = 0.185\linewidth]{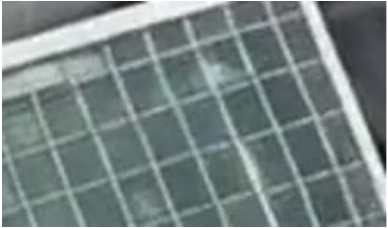} \\
(e) Base$_{\text{w/ FF\&ACFM}}$  &\hspace{-0.5mm}  
(f) $\text{Ours}_{\text{w/ ITE}}$  &\hspace{-0.5mm} 
(g) $\text{Ours}_{\text{w/o FF}\&\text{w/ US}}$  &\hspace{-0.5mm}   
(h) Ours \\
\end{tabular}}
\vspace{-3mm}
\caption{Effectiveness of the anchor-frame guided enhancement.}
\label{fig: effectiveness_Dit_1}
\vspace{-7.5mm}
\end{figure}

To further analyze the effect of the anchor-corresponding feature modulation (ACFM) module, we compare with a baseline method that integrates the ACFM to \emph{Base$_{\text{w/ FF}}$} (\emph{Base$_{\text{w/ FF}\&\text{ACFM}}$} for short).
As shown in Table~\ref{tab: ablation_study_all}, the addition of ACFM yields a substantial performance gain, increasing MUSIQ and DOVER by $2.66$ and $1.31$, respectively.
The visual comparison in Figure~\ref{fig: effectiveness_Dit_1}(c) and (e) shows that ACFM effectively strengthens the structural information (e.g., the grids in Figure~\ref{fig: effectiveness_Dit_1}(e)).
In addition, we compare with a baseline that replaces ACFM with a $3 \times 3$ depth-wise convolution (\emph{Base$_{\text{w/ FF}\&\text{DWC}}$} for short).
Both the quantitative result in Table~\ref{tab: ablation_study_all} and the qualitative comparison in Figure~\ref{fig: effectiveness_Dit_1} consistently validate the effectiveness of ACFM, demonstrating its capacity for discriminative local spatial feature selection and integration to facilitate VSR.

We additionally investigate the necessity of the anchor-frame feature refinement module.
As Table~\ref{tab: ablation_study_all} shows, our full model with the anchor-frame refinement achieves better performance, cf. $73.57/60.81$ for \emph{Ours} \emph{vs.} $73.24/59.99$ for \emph{Base$_{\text{w/ FF}\&\text{ACFM}}$}.
Figure~\ref{fig: effectiveness_Dit_1} shows that our method produces significantly clearer structural details.

Furthermore, we explore whether incorporating the anchor-frame features into the cross-attention layer in each DiT block (\emph{Ours$_{\text{w/ ITE}}$} for short) can generate better results.
Table~\ref{tab: ablation_study_all} shows that the interaction between the anchor-frame features and text embeddings compromises the structural information inherent in the anchor-frame features, which decreases MUSIQ by $4.84$ and thus leads to inferior visual quality in Figure~\ref{fig: effectiveness_Dit_1}(f).

Finally, to validate our anchor-frame selection strategy, we compare the proposed method with a baseline that employs uniform sampling (\emph{Ours$_{\text{w/o FF}\&\text{w/ US}}$} for short). 
The comparison results in Table~\ref{tab: ablation_study_all} and Figure~\ref{fig: effectiveness_Dit_1} show the superiority of our approach, demonstrating that the first frame, unaffected by temporal compression, provides richer structural information that is more effective for detail restoration.

\section{Conclusion}

We have presented STCDiT, an effective framework built upon a pre-trained video diffusion model for VSR, which restores structurally faithful while temporally stable video content from degraded inputs.
Compared with the standard VAE reconstruction process, our proposed motion-aware VAE reconstruction approach is more effective for reconstructing videos that exhibit various camera motion scenarios.
We further develop an anchor-frame guidance approach that leverages the rich structural information inherent in the first-frame latent extracted by the VAE encoder in each clip, termed the anchor-frame latent, to constrain the generation process toward faithful VSR.
Coupling these two designs allows our diffusion model to achieve high-quality video restoration.
Extensive evaluations and comparisons with state-of-the-art methods demonstrate that our approach achieves better performance in VSR.

{
    \small
    \bibliographystyle{ieeenat_fullname}
    \bibliography{main}

@String(CVPR= {IEEE Conf. Comput. Vis. Pattern Recog.})

@String(ICCV= {Int. Conf. Comput. Vis.})

@String(ECCV= {Eur. Conf. Comput. Vis.})

@String(NIPS= {Adv. Neural Inform. Process. Syst.})

@String(ICLR = {Int. Conf. Learn. Represent.})

@String(IJCAI = {IJCAI})

@String(AAAI = {AAAI})

@String(SPL	= {IEEE Sign. Process. Letters})

@String(CVPR  = {CVPR})

@String(ICCV  = {ICCV})

@String(ECCV  = {ECCV})

@String(NIPS  = {NeurIPS})

@String(ICLR  = {ICLR})

@inproceedings{PASD,
  title={Pixel-aware stable diffusion for realistic image super-resolution and personalized stylization},
  author={Yang, Tao and Wu, Rongyuan and Ren, Peiran and Xie, Xuansong and Zhang, Lei},
  booktitle={ECCV},
  year={2024}
}

@inproceedings{SeeSR,
  title={Seesr: Towards semantics-aware real-world image super-resolution},
  author={Wu, Rongyuan and Yang, Tao and Sun, Lingchen and Zhang, Zhengqiang and Li, Shuai and Zhang, Lei},
  booktitle={CVPR},
  year={2024}
}

@inproceedings{Dit4SR,
  title={DiT4SR: Taming Diffusion Transformer for Real-World Image Super-Resolution},
  author={Duan, Zheng-Peng and Zhang, Jiawei and Jin, Xin and Zhang, Ziheng and Xiong, Zheng and Zou, Dongqing and Ren, Jimmy and Guo, Chun-Le and Li, Chongyi},
  booktitle={ICCV},
  year={2025}
}

@inproceedings{DiffVSR,
  title={DiffVSR: Revealing an Effective Recipe for Taming Robust Video Super-Resolution Against Complex Degradations},
  author={Li, Xiaohui and Liu, Yihao and Cao, Shuo and Chen, Ziyan and Zhuang, Shaobin and Chen, Xiangyu and He, Yinan and Wang, Yi and Qiao, Yu},
  booktitle={ICCV},
  year={2025}
}

@inproceedings{UAV,
  title={Upscale-a-video: Temporal-consistent diffusion model for real-world video super-resolution},
  author={Zhou, Shangchen and Yang, Peiqing and Wang, Jianyi and Luo, Yihang and Loy, Chen Change},
  booktitle={CVPR},
  year={2024}
}

@inproceedings{MGLD,
  title={Motion-guided latent diffusion for temporally consistent real-world video super-resolution},
  author={Xi Yang and Chenhang He and Jianqi Ma and Lei Zhang},
  booktitle={ECCV},
  year={2024}
}

@inproceedings{Cogvideox,
  title={Cogvideox: Text-to-video diffusion models with an expert transformer},
  author={Yang, Zhuoyi and Teng, Jiayan and Zheng, Wendi and Ding, Ming and Huang, Shiyu and Xu, Jiazheng and Yang, Yuanming and Hong, Wenyi and Zhang, Xiaohan and Feng, Guanyu and others},
  booktitle={ICLR},
  year={2025}
}

@article{Wan,
  title={Wan: Open and advanced large-scale video generative models},
  author={Wan, Team and Wang, Ang and Ai, Baole and Wen, Bin and Mao, Chaojie and Xie, Chen-Wei and Chen, Di and Yu, Feiwu and Zhao, Haiming and Yang, Jianxiao and others},
  journal={arXiv preprint arXiv:2503.20314},
  year={2025}
}

@article{UltraVideo,
  title={UltraVideo: High-Quality UHD Video Dataset with Comprehensive Captions},
  author={Xue, Zhucun and Zhang, Jiangning and Hu, Teng and He, Haoyang and Chen, Yinan and Cai, Yuxuan and Wang, Yabiao and Wang, Chengjie and Liu, Yong and Li, Xiangtai and others},
  journal={arXiv preprint arXiv:2506.13691},
  year={2025}
}

@inproceedings{LSDIR,
  title={Lsdir: A large scale dataset for image restoration},
  author={Li, Yawei and Zhang, Kai and Liang, Jingyun and Cao, Jiezhang and Liu, Ce and Gong, Rui and Zhang, Yulun and Tang, Hao and Liu, Yun and Demandolx, Denis and others},
  booktitle={CVPR Workshops},
  year={2023}
}

@inproceedings{RealBasicVSR,
  title={Investigating tradeoffs in real-world video super-resolution},
  author={Chan, Kelvin CK and Zhou, Shangchen and Xu, Xiangyu and Loy, Chen Change},
  booktitle={CVPR},
  year={2022}
}

@article{QwenVL,
  title={Qwen2. 5-vl technical report},
  author={Bai, Shuai and Chen, Keqin and Liu, Xuejing and Wang, Jialin and Ge, Wenbin and Song, Sibo and Dang, Kai and Wang, Peng and Wang, Shijie and Tang, Jun and others},
  journal={arXiv preprint arXiv:2502.13923},
  year={2025}
}

@inproceedings{RealVSR,
  title={Real-world video super-resolution: A benchmark dataset and a decomposition based learning scheme},
  author={Yang, Xi and Xiang, Wangmeng and Zeng, Hui and Zhang, Lei},
  booktitle={ICCV},
  year={2021}
}

@InProceedings{REDS,
  author = {Nah, Seungjun and Baik, Sungyong and Hong, Seokil and Moon, Gyeongsik and Son, Sanghyun and Timofte, Radu and Lee, Kyoung Mu},
  title = {NTIRE 2019 Challenge on Video Deblurring and Super-Resolution: Dataset and Study},
  booktitle = {CVPR Workshops},
  year = {2019}
}

@inproceedings{UDM10,
  title={Detail-revealing deep video super-resolution},
  author={Tao, Xin and Gao, Hongyun and Liao, Renjie and Wang, Jue and Jia, Jiaya},
  booktitle={ICCV},
  year={2017}
}

@inproceedings{Flow_matching,
  title={Scaling rectified flow transformers for high-resolution image synthesis},
  author={Esser, Patrick and Kulal, Sumith and Blattmann, Andreas and Entezari, Rahim and M{\"u}ller, Jonas and Saini, Harry and Levi, Yam and Lorenz, Dominik and Sauer, Axel and Boesel, Frederic and others},
  booktitle={ICML},
  year={2024}
}

@article{VividVR,
  title={Vivid-VR: Distilling Concepts from Text-to-Video Diffusion Transformer for Photorealistic Video Restoration},
  author={Bai, Haoran and Chen, Xiaoxu and Yang, Canqian and He, Zongyao and Deng, Sibin and Chen, Ying},
  journal={arXiv preprint arXiv:2508.14483},
  year={2025}
}

@inproceedings{SUPIR,
  title={Scaling up to excellence: Practicing model scaling for photo-realistic image restoration in the wild},
  author={Yu, Fanghua and Gu, Jinjin and Li, Zheyuan and Hu, Jinfan and Kong, Xiangtao and Wang, Xintao and He, Jingwen and Qiao, Yu and Dong, Chao},
  booktitle={CVPR},
  year={2024}
}

@inproceedings{Star,
  title={Star: Spatial-temporal augmentation with text-to-video models for real-world video super-resolution},
  author={Xie, Rui and Liu, Yinhong and Zhou, Penghao and Zhao, Chen and Zhou, Jun and Zhang, Kai and Zhang, Zhenyu and Yang, Jian and Yang, Zhenheng and Tai, Ying},
  booktitle={ICCV},
  year={2025}
}

@inproceedings{Seedvr,
  title={Seedvr: Seeding infinity in diffusion transformer towards generic video restoration},
  author={Wang, Jianyi and Lin, Zhijie and Wei, Meng and Zhao, Yang and Yang, Ceyuan and Loy, Chen Change and Jiang, Lu},
  booktitle={CVPR},
  year={2025}
}

@article{Seedvr2,
  title={Seedvr2: One-step video restoration via diffusion adversarial post-training},
  author={Wang, Jianyi and Lin, Shanchuan and Lin, Zhijie and Ren, Yuxi and Wei, Meng and Yue, Zongsheng and Zhou, Shangchen and Chen, Hao and Zhao, Yang and Yang, Ceyuan and others},
  journal={arXiv preprint arXiv:2506.05301},
  year={2025}
}

@inproceedings{DOVE,
  title={DOVE: Efficient One-Step Diffusion Model for Real-World Video Super-Resolution},
  author={Chen, Zheng and Zou, Zichen and Zhang, Kewei and Su, Xiongfei and Yuan, Xin and Guo, Yong and Zhang, Yulun},
  booktitle={NeurIPS},
  year={2025}
}

@inproceedings{window_based_1,
  title={Real-time video super-resolution with spatio-temporal networks and motion compensation},
  author={Caballero, Jose and Ledig, Christian and Aitken, Andrew and Acosta, Alejandro and Totz, Johannes and Wang, Zehan and Shi, Wenzhe},
  booktitle={CVPR},
  year={2017}
}

@article{window_based_2,
  title={Video super-resolution via bidirectional recurrent convolutional networks},
  author={Huang, Yan and Wang, Wei and Wang, Liang},
  journal={IEEE TPAMI},
  year={2017},
}

@inproceedings{window_based_3,
  title={Video super-resolution with temporal group attention},
  author={Isobe, Takashi and Li, Songjiang and Jia, Xu and Yuan, Shanxin and Slabaugh, Gregory and Xu, Chunjing and Li, Ya-Li and Wang, Shengjin and Tian, Qi},
  booktitle={CVPR},
  year={2020}
}

@inproceedings{window_based_4,
  title={Mucan: Multi-correspondence aggregation network for video super-resolution},
  author={Li, Wenbo and Tao, Xin and Guo, Taian and Qi, Lu and Lu, Jiangbo and Jia, Jiaya},
  booktitle={ECCV},
  year={2020},
}

@inproceedings{basicvsr++,
  title={Basicvsr++: Improving video super-resolution with enhanced propagation and alignment},
  author={Chan, Kelvin CK and Zhou, Shangchen and Xu, Xiangyu and Loy, Chen Change},
  booktitle={CVPR},
  year={2022}
}

@inproceedings{basicvsr,
  title={Basicvsr: The search for essential components in video super-resolution and beyond},
  author={Chan, Kelvin CK and Wang, Xintao and Yu, Ke and Dong, Chao and Loy, Chen Change},
  booktitle={CVPR},
  year={2021}
}

@inproceedings{recurrent_based_1,
  title={Frame-recurrent video super-resolution},
  author={Sajjadi, Mehdi SM and Vemulapalli, Raviteja and Brown, Matthew},
  booktitle={CVPR},
  year={2018}
}

@inproceedings{recurrent_based_2,
  title={Bidirectional recurrent convolutional networks for multi-frame super-resolution},
  author={Huang, Yan and Wang, Wei and Wang, Liang},
  booktitle={NIPS},
  year={2015}
}

@article{recurrent_based_3,
  title={Collaborative feedback discriminative propagation for video super-resolution},
  author={Li, Hao and Chen, Xiang and Dong, Jiangxin and Tang, Jinhui and Pan, Jinshan},
  journal={arXiv preprint arXiv:2404.04745},
  year={2024}
}

@article{recurrent_based_7,
  title={Cascaded Temporal Updating Network for Efficient Video Super-Resolution},
  author={Li, Hao and Dong, Jiangxin and Pan, Jinshan},
  journal={arXiv preprint arXiv:2408.14244},
  year={2024}
}

@article{recurrent_based_5,
  title={Rethinking alignment in video super-resolution transformers},
  author={Shi, Shuwei and Gu, Jinjin and Xie, Liangbin and Wang, Xintao and Yang, Yujiu and Dong, Chao},
  journal={NIPS},
  year={2022}
}

@inproceedings{recurrent_based_6,
  title={Video super-resolution transformer with masked inter\&intra-frame attention},
  author={Zhou, Xingyu and Zhang, Leheng and Zhao, Xiaorui and Wang, Keze and Li, Leida and Gu, Shuhang},
  booktitle={CVPR},
  year={2024}
}

@inproceedings{GAN_based_1,
  title={NegVSR: Augmenting negatives for generalized noise modeling in real-world video super-resolution},
  author={Song, Yexing and Wang, Meilin and Yang, Zhijing and Xian, Xiaoyu and Shi, Yukai},
  booktitle={AAAI},
  year={2024}
}

@inproceedings{GAN_based_2,
  title={Realviformer: Investigating attention for real-world video super-resolution},
  author={Zhang, Yuehan and Yao, Angela},
  booktitle={ECCV},
  year={2024}
}

@inproceedings{GAN_based_3,
  title={Videogigagan: Towards detail-rich video super-resolution},
  author={Xu, Yiran and Park, Taesung and Zhang, Richard and Zhou, Yang and Shechtman, Eli and Liu, Feng and Huang, Jia-Bin and Liu, Difan},
  booktitle={CVPR},
  year={2025}
}

@inproceedings{CLIPIQA,
  title={Exploring clip for assessing the look and feel of images},
  author={Wang, Jianyi and Chan, Kelvin CK and Loy, Chen Change},
  booktitle={AAAI},
  year={2023}
}

@inproceedings{MUSIQ,
  title={Musiq: Multi-scale image quality transformer},
  author={Ke, Junjie and Wang, Qifei and Wang, Yilin and Milanfar, Peyman and Yang, Feng},
  booktitle={CVPR},
  year={2021}
}

@inproceedings{Lpips,
  title={The unreasonable effectiveness of deep features as a perceptual metric},
  author={Zhang, Richard and Isola, Phillip and Efros, Alexei A and Shechtman, Eli and Wang, Oliver},
  booktitle={CVPR},
  year={2018}
}

@inproceedings{MANIQA,
  title={Maniqa: Multi-dimension attention network for no-reference image quality assessment},
  author={Yang, Sidi and Wu, Tianhe and Shi, Shuwei and Lao, Shanshan and Gong, Yuan and Cao, Mingdeng and Wang, Jiahao and Yang, Yujiu},
  booktitle={CVPR},
  year={2022}
}

@article{NIQE,
  title={Making a “completely blind” image quality analyzer},
  author={Mittal, Anish and Soundararajan, Rajiv and Bovik, Alan C},
  journal={IEEE SPL},
  year={2012}
}

@inproceedings{DOVER,
  title={Exploring video quality assessment on user generated contents from aesthetic and technical perspectives},
  author={Wu, Haoning and Zhang, Erli and Liao, Liang and Chen, Chaofeng and Hou, Jingwen and Wang, Annan and Sun, Wenxiu and Yan, Qiong and Lin, Weisi},
  booktitle={ICCV},
  year={2023}
}

@inproceedings{faithdiff,
  title={Faithdiff: Unleashing diffusion priors for faithful image super-resolution},
  author={Chen, Junyang and Pan, Jinshan and Dong, Jiangxin},
  booktitle={CVPR},
  year={2025}
}

@inproceedings{LIQE,
  title={Blind image quality assessment via vision-language correspondence: A multitask learning perspective},
  author={Zhang, Weixia and Zhai, Guangtao and Wei, Ying and Yang, Xiaokang and Ma, Kede},
  booktitle={CVPR},
  year={2023}
}

@article{FasterVQA,
  title={Neighbourhood representative sampling for efficient end-to-end video quality assessment},
  author={Wu, Haoning and Chen, Chaofeng and Liao, Liang and Hou, Jingwen and Sun, Wenxiu and Yan, Qiong and Gu, Jinwei and Lin, Weisi},
  journal={IEEE TPAMI},
  year={2023}
}

@inproceedings{shi1994good,
  title={Good features to track},
  author={Shi, Jianbo and others},
  booktitle={CVPR},
  year={1994}
}

@inproceedings{Real-ESRGAN,
  title={Real-esrgan: Training real-world blind super-resolution with pure synthetic data},
  author={Wang, Xintao and Xie, Liangbin and Dong, Chao and Shan, Ying},
  booktitle={CVPR},
  year={2021}
}

@inproceedings{omnissr,
  title={Omnissr: Zero-shot omnidirectional image super-resolution using stable diffusion model},
  author={Li, Runyi and Sheng, Xuhan and Li, Weiqi and Zhang, Jian},
  booktitle={ECCV},
  year={2024}
}

@inproceedings{lucas1981iterative,
  title={An iterative image registration technique with an application to stereo vision},
  author={Lucas, Bruce D and Kanade, Takeo},
  booktitle={IJCAI},
  year={1981}
}

@article{su2024roformer,
  title={Roformer: Enhanced transformer with rotary position embedding},
  author={Su, Jianlin and Ahmed, Murtadha and Lu, Yu and Pan, Shengfeng and Bo, Wen and Liu, Yunfeng},
  journal={Neurocomputing},
  year={2024},
}

@inproceedings{Lora_1,
  title={Towards Higher Effective Rank in Parameter-Efficient Fine-tuning using Khatri-Rao Product},
  author={Albert, Paul and Zhang, Frederic Z and Saratchandran, Hemanth and van den Hengel, Anton and Abbasnejad, Ehsan},
  booktitle={ICCV},
  year={2025}
}

@inproceedings{Lora_2,
  title={RandLoRA: Full-rank parameter-efficient fine-tuning of large models},
  author={Albert, Paul and Zhang, Frederic Z and Saratchandran, Hemanth and Rodriguez-Opazo, Cristian and Hengel, Anton van den and Abbasnejad, Ehsan},
  booktitle={ICLR},
  year={2025}
}

@inproceedings{Lora_3,
  title={Efficient learning with sine-activated low-rank matrices},
  author={Ji, Yiping and Saratchandran, Hemanth and Gordon, Cameron and Zhang, Zeyu and Lucey, Simon},
  booktitle={ICLR},
  year={2025}
}
}

\end{document}